# Group Activity Recognition in Computer Vision: A Comprehensive Review, Challenges, and Future Perspectives

**CHUANCHUAN WANG[1,2], AHMAD SUFRIL AZLAN MOHAMED [1, *]**

[1] School of Computer Science, Universiti Sains Malaysia, Penang, 11800, Malaysia
[2] School of Engineering, College of Technology and Business, Guangzhou, 510850, China

*Corresponding author: Ahmad Sufril Azlan Mohamed (sufril@usm.my)

**Abstract:** Group activity recognition is a hot topic in computer vision. Recognizing activities through group relationships plays a vital role in group activity recognition. It holds practical implications in various scenarios, such as video analysis, surveillance, automatic driving, and understanding social activities. The model's key capabilities encompass efficiently modeling hierarchical relationships within a scene and accurately extracting distinctive spatiotemporal features from groups. Given this technology's extensive applicability, identifying group activities has garnered significant research attention. This work examines the current progress in technology for recognizing group activities, with a specific focus on global interactivity and activities. Firstly, we comprehensively review the pertinent literature and various group activity recognition approaches, from traditional methodologies to the latest methods based on spatial structure, descriptors, non-deep learning, hierarchical recurrent neural networks (HRNN), relationship models, and attention mechanisms. Subsequently, we present the relational network and relational architectures for each module. Thirdly, we investigate methods for recognizing group activity and compare their performance with state-of-the-art technologies. We summarize the existing challenges and provide comprehensive guidance for newcomers to understand group activity recognition. Furthermore, we review emerging perspectives in group activity recognition to explore new directions and possibilities.

## 1 Introduction

As an indispensable branch of artificial intelligence, computer vision empowers computational models to extract valuable and structured high-level semantic information from raw image data[1]. In the early stages of computer vision research, the focus was on image feature extraction. However, with the advancements in sensor and camera technology and expanding knowledge, the emphasis has shifted toward studying methods for activity recognition based on videos. The increasing availability of video data has led to a growing demand for video understanding technology in real-life applications. For instance, security is crucial in identifying abnormal activities and providing real-time warnings. On the internet, video understanding helps extract structured information for downstream models, such as content interactivity and recommendation.

Consequently, the problem of visual comprehension with video data as input and developing video comprehension technology have emerged as prominent research directions[2]. Videos introduce a temporal dimension, necessitating the modeling of multi-frame information to understand dynamic time sequences. However, challenges arise from irrelevant redundancy and motion blur, leading to computational overhead. Thus, processing and comprehending video data pose significant challenges.

This work focuses on group activity recognition in videos, where the model aims to classify ongoing activities in the video involving multiple individuals, such as hugging, handling, playing volleyball, etc. Previous research on video activity recognition was primarily confined to single-person scenarios [3], where videos featured only one character[4]. However, through further investigations[5, 6], a task for recognizing group activities in multi-person scenes has been proposed and explored. In this joint activity recognition task, the input video contains multiple individuals, and the model is required to identify collective activities that involve various people performing actions together, such as walking, running, and playing as a group. Figure 1 illustrates an example of joint activity recognition, where in video screen 6, multiple volleyball players collaborate to execute the collective activities of "left attack" and "right defending."

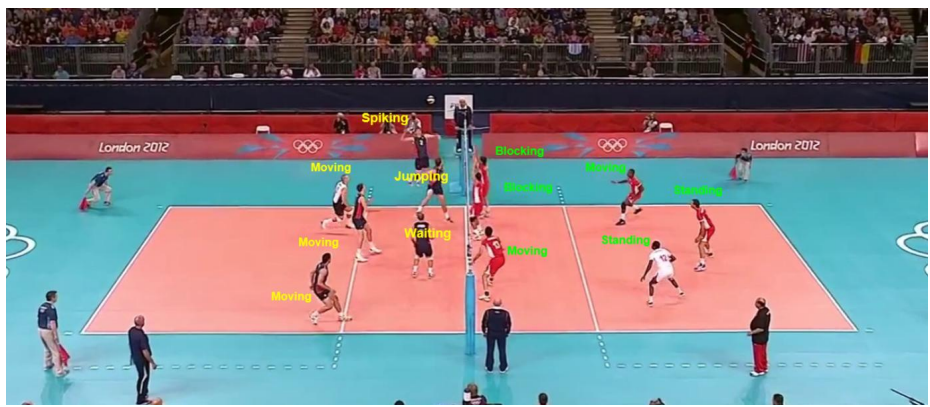

Figure 1 Volleyball group activity of "left attack" and "right defending" 6

Group activity recognition empowers video models to address complex, multiplayer scenes and holds great promise for various applications. Security monitoring facilitates recognizing group activities and alerts to abnormal behaviors. In the medical domain, it finds applications in fall detection and enables identifying and analyzing athletes' movements in sports competitions. Given its wide-ranging applications, group activity recognition has attracted considerable attention in research. Moreover, numerous other application scenarios exist. This paper presents a comprehensive analysis of recent research and challenges in group activity recognition, summarizes group activity recognition methods, and outlines future research directions.

Compared with the related surveys, this work focuses on a comprehensive investigation of interactivity relational human activity recognition, the application progress of group activity recognition, and the accuracy comparison of state-of-the-technologies on public benchmarks. We investigated the related reviews and their main contents, as shown in Table 1, which includes citations based on Google Scholar or Web of Science and their impact factor (IF) value. This review aims to comprehensively understand group activity recognition, its relational network, mechanism, classification, and application progress. Moreover, it provides an experimental comparison for the investigated-on assessment analysis. Meanwhile, it mainly provides developing prospects or new ideas for human activity recognition researchers.

Table 1 The related surveys and their main contents are referenced in this work.

| Paper | Year | description | Classification | Citation | IF |
|---|---|---|---|---|---|
| [7] | 2008 | Survey of human activity and Activity recognition from video | Activities and activities | 1791 | 5.86 |
| [8] | 2013 | A description for activity recognition in video surveillance | Video surveillance | 443 | 2.84 |
| [9] | 2014 | The background knowledge and available features are relevant to analyzing crowded activity scenes introduced. | scene analysis | 494 | 5.86 |
| [10] | 2015 | Present methods exploit these semantic features to recognize activities from still images, video data, and four groups of activities. | semantic features | 234 | 8.52 |
| [11] | 2018 | Review current CNN methods that outperform earlier methods using hand-crafted representations. | CNN-based approaches | 523 | 4.76 |
| [12] | 2019 | Survey of deep learning for crowd video analysis and violence detection. | video surveillance | 107 | 10.84 |
| [13] | 2020 | Review and comparison of crowd analysis methods, datasets, and taxonomies. | Data acquired from various types | 269 | 8.52 |
| [14] | 2022 | Review of deep learning for crowd analysis: classification, datasets, methods, and applications. | Pixel level, texture level, and object level | 8 | 8.17 |
| [15] | 2023 | Review recent advances in deep learning methods for GAR and categorize them by the type of input data modality. | Modalities | 19 | 24.31 |

The main contributions of this investigation are the following:

a. This systematic review is the first to classify technologies for recognizing group-related activities into six approaches. As illustrated in the taxonomy, it provides a clear introduction to different state-of-the-art technologies, establishing a significant milestone in the field.

b. The interactivity relational network model has been reviewed, and the attention mechanism, such as group-attentional models, are also studied, which play a crucial role in group activity recognition.

c. We give a detailed description of popular datasets used to analyze the algorithms' performance evaluation and briefly introduce other relative categories of benchmark data sets.

d. Existing problems and developing prospects of interactivity relation activity recognition are pointed out, which gives future directions to the beginner or subsequent researcher.

We have organized the remaining sections of this paper as follows. Section 2 introduces six categories of interactivity activity recognition methods from traditional to deep learning approaches. Section 3 presents several popular group activity recognition datasets used by researchers, brief taxonomy of benchmark datasets, and compares the accuracy of state-of-the-technologies methods. Then, Section 4 discusses the possible future research directions on the problem of group activity recognition. Finally, we summarize the entire work in Section 5.

## 2. Group Activity Recognition

Group activity recognition is a sub-field from group activity, which involves identifying activities performed by multiple individuals in a video with a multi-person scene. Current existing approaches dominated by neural network methods, such as CNN-based, GNN-based, RNN-based, GCN-based, and LSTM-based, often leverage consists of complicated architectures and mechanisms to embed the relationships between the two persons on the architecture[16]. This section classifies six group relation activity methods based on spatial structure, descriptor, non-deep learning, Hierarchical Recurrent Neural Networks (HRNN), relationship model, and attention mechanism. We introduce these methods from the perspective of traditional to the latest approaches. We arrange these methods in descending order by year. Table 2 presents the taxonomy of this review, showcasing a selection of algorithms for recognizing group-related activities.

Table 2 **Taxonomy of this review and the lists of selected methods with year**

| Spatial structure-based | DCIO[17](2003);STRM[18](2009);NUS-HGA[19](2009); GPE[20](2010) |
|---|---|
| Descriptor-based | STR[21](2009);AC[22](2010);RSTV[23](2011); interactive phrases[24] (2012)VIAR[25](2012) |
| Non-deep Learning-based | MIR[26](2015);CK[27](2015);DCM[28](2017) |
| HNN-based Methods | HDTM[5](2016);CERN[29](2017);SSU[30](2017); DS-LSTM[31](2019) |
| Relationship Model-based | stageNet[32](2018);HRN[33](2018);DR$^2$N[34](2019)ARG[35](2019);IRN[16](2021);STIPN[36](2021);Dual-AI[37] (2022) ;GIRN[38](2022);STIGPN[39](2023) |
| Attention-based | HANs+HCN[40](2018);SPA[41](2018);GAIM[42](2019);AT[43](2020);RGB+Depth[44](2022) |

**2.1 Spatial Structure-based Methods**

Interactivity activity refers to the group activity between multiple individuals in a scene. In the early stages of the study, researchers leverage target-tracking algorithms or other sensors to obtain the location of each character and group activity based on the spatial location structure. Researchers have proposed various methods to extract features from the movement trajectory of characters and represent the population as a whole for identifying group activity. These methods include shape-based modeling, three-level local motor relationships, a Gaussian process approach, and a probabilistic connectivity map. These approaches leverage different feature descriptors and classifiers to identify group activity, such as walk-in-group, run-in-group, stand-and-talk, gathering, fighting, and ignoring, as shown in Figure 2. In complex scenes, there may be multiple groups of characters, and different group activities are being the way, requiring the model to divide the population and analyze group activity separately for each small group.

Initial studies on group behavior recognition focused on spatial crowd organization to identify individuals based on their location and motion patterns, straightforward group activities like gathering, catching up, dispersing, and walking together. The method treats the video as a 2D plane, considering people as moving points and extracting their characteristics for classification. However, this approach has limitations in distinguishing complex group behaviors unrelated to location.

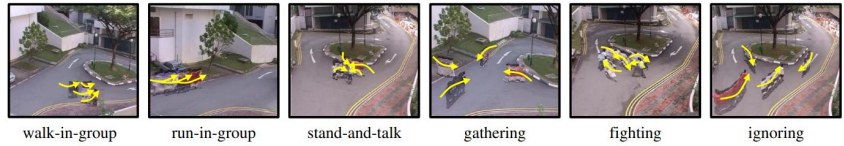

Figure 2 An illustration of Spatial Local-based Interactivity Activity Recognition[45]

To model the position of all characters at each moment as a polygon and to determine whether abnormal activity occurs in the current video, Vaswani et al.[17] proposed shape-based modeling can view the entire population as a whole rather than modeling each individual individually. The authors leverage the Kendall shape theory[46] to describe polygonal shapes for each frame and estimate regular activity shape probability distributions.

Khan & Shah proposed a method based on the overall modeling of multiple characters in the spatial structure of the population[47]. The author first represents the character formation into three-dimensional polygons, where each vertex is the individual character in the crowd, and the motion-tracking track of nature is taken as the feature of each vertex so that the group shape-tracking matrix can obtain. Then, researchers decompose the matrix, estimate its rank, and ultimately classify the group activity of the population.

The dynamic relationship of character movement trajectory is significant for group activity identification. Ni et al.[19] proposed to leverage three levels of local motor relationships to encode group activity, including self, pairwise, and group relationships. Specifically, the self-relationship represents the internal connection of each character's movement trajectory and how each moment changes compared to the last moment, depicting the activity of the individual personality.

To classify interactivity activity, Cheng et al.[48] Using a motion trajectory analysis method, he introduced a Gaussian process approach to represent the motion trajectories, handling the randomness of the motion from a probabilistic perspective. Based on Gaussian process regression and motion analysis, Cheng et al. designed three feature descriptors to extract the features of the motion trajectory: individual features, paired features, and overall features. Finally, based on the feature word bag method, the population with unfixed numbers is represented as fixed-length feature vectors, and the group activity was classified using the SVM.

Y. Zhang et al.[49] proposed a method to extract and classify motion trajectory features for group activity recognition using probabilistic mutating and four group feature extractive methods. Some complex scenes may have multiple characters and different group activities. In this case, the model needs to have the ability to divide the population, locate the distribution of each small group, and combine it with the relationship between multiple small groups to analyze the group activity of each small group separately. Yin et al.[50]first proposed a method to recognize group efforts in complex scenes with multiple characters and groups. The technique uses a multi-object tracking algorithm and a minimum spanning tree algorithm to cluster the symbols into groups and track their motion trajectories. The method also uses a subset of feature descriptors based on social network analysis to extract a feature vector for each group and a Gaussian process dynamic model to classify the group activity.

C. Zhang et al.[51] analyzed the hierarchy of personality interactivity and proposed that personality interactivity activity includes three levels: individual personality, subpersonality, and overall personality interactivity. These three levels of Relational information are essential for identifying complex interactivities, and previous work has neglected relational communication between sub-interactivity. Based on these analyses, C. Zhang et al. refer to the relationship between suboperations in the group activity recognition framework, which considers the interrelationship between individual characters and extracts the interactivity information between multiple sub-groups. Azorin Lopez et al. [52] extended the descriptive activity vectors that perform well on the single-person activity recognition task. They propose a novel group activity illustrative vector based on character movement trajectories to identify descriptors for interactive activities. It extracts the group activity characteristics from three aspects: the group movement trajectory, the correlation of the individual movement trajectory, and the group movement relationship of different sub-groups.

Identifying interactivity activity to solve the problem of a large amount of noise information in complex multiplayer scenes is challenging. Cho et al.[53] proposed the concept of group interactivity spatial. Spatial interactivity pertains to the grouping of individuals actively participating in a collective activity within a given scene. Cho et al. [53] introduced a novel spatial detection method for identifying group interactivity by leveraging spatial relations. Within the specified population interactivity space, the authors presented two innovative feature descriptors to characterize the population's activities effectively: population interactivity energy features and gravitational repulsion features. These descriptors consider the positional relationships between the characters, including proximity, distance, and maintaining distance. Finally, the authors utilized the feature word bag method and linear SVM to accurately classify group activity.

**2.2 Descriptor-based Methods**

In addition to spatial structure information, visual appearance information is critical for identifying interactivity activity. Early algorithms used hand-designed feature descriptors, such as Histogram of Oriented Gradients(HOG)[54], to extract visual features. At the same time, more recent methods involve using multivariate feature descriptors and random forest-based algorithms to learn contextual features automatically. The proposed algorithms include tracking characters' movements, designing local spatiotemporal descriptors, and using fully connected conditional random field models to reduce local classification errors.

To achieve robust recognition in the interactivity scene, Choi et al.[6] simultaneously use the spatial distribution of people, the posture of people, and the movement of people to identify the group activity in the scene. The proposed solution first uses an algorithm based on the HOG descriptor to detect characters in the video frame and estimate their pose. Then, it uses an extended Kalman filter to track the movements of all characters in the scene. Based on the previous collective correlation idea, Choi et al.[23] proposed a random forest-based algorithm to allow the model to automatically learn the contextual features of the group for group activity recognition. Similarly, interactive phrases[24] semantic description of interactivity activity recognition approach can express binary semantic motion relationships between interacting people.

The vast majority of group activity research work focuses on the problem of activity recognition, where the model classifies a given video. However, a random forest-based algorithm [23] proposes the task of activity retrieval. The model needs to find the person most related to the activity label from the video. For some activities with shallow frequency, such as abnormal activities such as falling and fighting, there is a large gap between positive and negative samples, and it is challenging to train the corresponding classifier to identify this activity. In addition, the author also designed the character descriptor and the activity context descriptor to extract the characteristics of each character and its surrounding environment.

Since the activity context descriptor[23] is highly sensitive to perspective changes and lacks robustness, it poses limitations. Kaneko et al.[55] proposed a new, improved feature descriptor: relative activity context(RAC) descriptor to encode the close relationship in the group, which has the advantage of perspective invariance. The RAC descriptors also put forward two other methods to improve model performance. First, to make the interactivity of context features robust, Kaneko et al. designed a threshold processing operation for group activity from the context of noise features and a Gaussian processing operation to reduce the characteristic value quantification error in the process. Moreover, the fully connected conditional random field model is leveraged in the classification process to reduce local classification errors.

Mohamed R Amer & Todorovic[56]presented a video feature that selects the local visual cues of the people involved in the target group activity from multi-person scenes. They also proposed a generative chain model that uses a maximal posterior probability algorithm to recognize interactivity. The algorithm iteratively optimizes the model and redefines the input feature subset to extract the optimal interactivity feature.

Nabi et al.[57]proposed a spatiotemporal feature descriptor based on pose semantics to simultaneously capture multiscale interactivities between multiple video characters. Character activity can regard as a specific sequence of character posture changes, so the critical parts of nature can detect in the video frame, such as head, body, and legs, and the character activity can judge based on the character posture information. Nabi et al. designed the temporal attitude descriptor, which first uses the attitude estimation algorithm to detect the position of the critical character on each video frame as the attitude semantic information. Then, they calculated the size distribution of activation values for all human classes at each spatiotemporal position to obtain the feature vector.

**2.3 Non-deep Learning-based Methods**

Group activity often contains many types of relationship information, such as the interactivity between people, the relationship between human and scene context, the relationship between individual and group, and the relationship between group and group. Modeling the hierarchical structure between the elements in the scene and using the dependencies between them can enable the model to better understand multiple complex activity scenes.

D. Zhang et al.[58] the group activity recognition problem was decomposed into individual and group levels, proposing a two-layer hidden Markov model framework to identify group activity in video sequences. Each layer can leverage different hidden Markov models for different sub-problems, making the whole model easy to scale. The multimodalities feature, encompassing audio and video, enables the model to identify group interactivity activities such as discussion, monologue, and speech, necessitating judgments based on sound and visuals. Dai et al.[59]an event-based dynamic context model proposed to address the context perception problem in a group interactivity recognition task. Dai et al. designed an event-driven multilevel dynamic Bayesian network to detect multilevel events in the scene. Lan et al.[60] leverage the hidden variable model framework to model the contextual information in the background, including character and group interactivity information and character and character interactivity information.

To address the problem of spatiotemporal consistency in group activity recognition, Kaneko et al. proposed a model based on fully connected conditional random fields for group activity identification[61]. This work establishes multiscale relationship connections between characters in the model from four aspects: spatial location, size, motion, and temporal position. To model the relationship between people in the group activity recognition problem, X. Chang et al.[26] proposed a learning-based method to compute the interactivity relationship between two characters and specify the activity category. Specifically, the model represents the interactive relationship between two characters by calculating the dot product similarity of the activity feature vectors of two atoms.

Mohamed R Amer et al.[62] proposed a 3-layer AND-OR graph model, which can hierarchically model multiple scenes, and simultaneously represent the participating individual objects, individual activities, and group activities in the background. Mohamed R Amer et al.[63] similarly leverage the AND-OR graph model to model hierarchical spatiotemporal relationships in a group activity. To further improve the inference efficiency of the model, Mohamed R Amer et al. proposed the AND-OR graph model inference algorithm based on a Monte Carlo tree search. The experimental results show that the speed of the AND-OR graph model inference algorithm based on the Monte Carlo tree search is two orders of magnitude compared to the previous algorithm[62, 64] while keeping the accuracy unchanged.

To capture the higher-order spatiotemporal dependence of long ranges between video features, 2014 Mohamed Rabie Amer et al.[65] proposed a hierarchical random field model. The background noise information is gradually filtered through layer-by-layer feature fusion to retain valuable group activity characteristics. Compared with other works that use conditional random fields to identify group activity, Mohamed Rabie Amer et al. adjusted the connection of hidden variables in the model, using two hidden layers to model the spatial association of single frames and the temporal correlation of multiple structures and separate the spatiotemporal features. Lan et al.[66] proposed the concept of social roles in group activity. For example, at some point in a sport, each player may act as an attacker, a defender, etc. Over time, each character changes. Social roles provide semantic information for the interactive relationships between characters.

Most group activity recognition methods assume that only one group activity occurs in the scene and that all the characters share a group activity context. Choi et al.[67] proposed that in actual application scenarios, multiple group activities will coincide, providing multiple contextual clues to each other. Therefore, Choi et al. proposed a holistic discriminative learning framework to identify group steps in scenarios based on numerous context models. First, they also consider the intra-and class interactivities in the population; Then, many techniques.

Due to the influence of the camera angle, the change in the character's appearance, and the timing of the activity, there will be many variations of the same category, which challenges the activity recognition algorithm. Lan et al.[68] proposed a method that efficiently models different activity variants within the same type to address this issue. They achieve this by discovering elements of activity classes and subcategories within each activity category.

Sometimes, we can treat many visual recognition problems as counting issues. For instance, when determining whether an event occurs in a long video, we need to measure the number of frames that contain the event; to determine whether a crowd is doing a group activity, you need to count the number of individual actions in the population. In these problems, making full use of the cardinality relationship between the child elements can reduce the model's sensitivity to the noise information, such as irrelevant video frames and characters in the population. Based on these ideas, Hajimirsadeghi et al. [27] proposed a flexible framework based on a multi-instance learning model to infer the base relationship between hidden labels and solve the ambiguity in label identification based on hard and soft base relations. Zhou et al.[69] proposed a generative model for the relationship modeling problem in the group activity recognition task. Zhou et al. designed a probabilistic graph model with four hierarchical structures: scene events, group activity, standard poses, and visible poses. Each character has four associated variables in the model, corresponding to these four levels. The model generates each variable top-down, from scene events to visible poses. Choi & Savarese[70] proposes a holistic framework for simultaneous multi-target person tracking and identifying their group activity. The proposed model can track multiple characters simultaneously, give each character's movement trajectory, and remember their atomic, interactivity, and group activities. Khamis et al.[71] used two contextual cues to improve the accuracy of group activity recognition. The first is the context of the individual character's movement trajectory, that is, the timing change information of each nature; The next aspect is the single-frame scene context, which includes the relational knowledge of the other characters in the scene. Tran et al.[72] uses undirected graphs to represent multiplayer scenes, where each node represents a character, and the weight of each edge indicates the degree of correlation between two characters. Sun et al.[73] devised a holistic framework to address the spatiotemporal group activity localization problem and proposed an implicit relational graph model to handle both tasks simultaneously, including multi-target tracking, joint mapping, and group activity recognition.

Multiple surveillance cameras must simultaneously identify group activity in various locations in some surveillance scenes, such as schools, parks, and factories. Group activity recognition faces significant challenges in multi-camera monitoring scenes due to many characters, complex settings, and character redundancy, posing enormous difficulties. Zha et al.[74] proposed a group activity detection method based on multi-camera context for multi-camera scenes, which can use the context information inside and between cameras simultaneously, and does not need to know the topology of multiple cameras. M.C. Chang et al.[75] also designed an abnormal activity detection algorithm for multi-camera scenes. M.C. Chang et al. first built a multi-target tracking system based on multiple cameras, then extracted each character's spatial and temporal characteristics based on the tracking results.

**2.4 HRNN-based Methods**

In recent years, deep learning technology based on the convolutional neural network has succeeded [76, 77] [78, 79] in various sub-tasks in computer vision, which has extensively promoted the performance of related models. On the problem of group activity recognition, many works have also begun to choose hierarchical recurrent neural networks to extract visual features in videos.

Mostafa S Ibrahim et al.[80] proposed a hierarchical deep temporal model two-stage LSTM, which uses long-term memory networks to model the temporal dynamics of the entire multiplayer scene. The model includes two stages: the long-short-term memory network in stage 1 and extracts the representation of the long-short-term memory network in stage 2. The authors employ a deep convolutional neural network to eliminate the visual characteristics of the characters.

Subsequently, the data undergoes a two-stage temporal model that progressively integrates low-level features into high-level information.

Wang et al.[81] proposes a recurrent interactivity context modeling framework for group activity understanding based on LSTM networks. The proposed group row uses three levels of LSTM networks to integrate the character characteristics layer by layer: the individual, group, and scene levels. Among them, the individual-level memory network is responsible for extracting representations to represent the dynamic changes of individual character movements. In contrast, the long-and short-term memory network at the group and scene levels integrates the character characteristics of the same group or the whole set, respectively. The three levels of networks can obtain different levels of context characteristics and can conduct multi-level multiplayer scene activity understanding. This framework adds group-level fusion networks compared to two-stage LSTM[80], which can handle complex scenarios containing multiple populations and provide interactive context features with greater description power. In addition, the authors used optical flow modes as model input and simultaneously extracted spatial and motion features using a dual-stream network for each character. T. Shu et al.[82] is also based on a two-stage long-and short-term memory network to extract scene features for group activity recognition. This approach optimizes the loss function of the group activity recognition model based on energy minimization and confidence maximization strategies. Li & Choo Chuah[83] proposed a semantic framework based on group activity recognition. Inspired by the success of recurrent neural networks in natural language processing and text generation, they use extended- and short-term memory networks to generate text for input videos and obtain the interpretable semantic text for semantic-based group activity recognition.

Some similar local movements may confuse the discrimination of different group activities. For example, in the two group activity categories of "gathering" and "separation," the actions of individual characters are both "walking." Group activities can be accurately identified only using contextual information and character relationships. P-S. Kim et al.[84] proposes a discriminative population context feature to enhance the recognition ability of the model, Which first identifies the subpopulations in the scene, then focuses on the relationships before the subpopulation, highlighting the differences before the activity of the different groups.

Group activity recognition models are generally trained based on supervised learning, requiring much-labeled data. Gammulle et al.[85] proposed a multi-sequence generation adversarial network that can perform semi-supervised learning for group activity recognition. Wu et al.[86] presented an Actor Relation Graph(ARG) to capture the appearance and position relation between actors simultaneously. To eliminate the noise interference of the complex background in the video, the authors use the optical flow to extract the global motion patterns in the video frame sequence and simultaneously use the spatial and temporal features of the worldwide motion patterns to identify the events in the video.

Based on the ARG [86], L. Wu et al.[87]optimized the extractive method of video movement mode, extracting the flow of the two groups of players and the flow of camera movement. In this work, the author analyzed the recognition of group activity in a basketball game and proposed that a semantic event in a basketball game often contains three stages: pre-event, event, and post-event. These three stages play a different role in the differentiation of scene events.

X. Shu et al.[88] proposed a model combining graph and long-term memory network to model the interdependence between individual and group activity. From the local perspective, the long-level short-term memory network extracts activity representation based on the interactivity between characters—the LSTM network models group activity based on graph structure in the global perspective. In addition, the author designed the residual long-term and short-term memory network to extract the temporal characteristics of the characters.

To identify the group activity in the video, it is often necessary to use the target detector to locate the position of all the characters in each video frame and then further extract the character features. Most work uses offline character detectors, removes good character candidate boxes in advance, and conducts group activity identification. Bagautdinov et al.[89] attempts to integrate the character detection and group activity recognition networks and train them end-to-end to form a complete multiplayer scene understanding model. To improve the accuracy of character detection, the author uses Markov random field to screen the credible candidate box from all the candidate boxes, replacing the traditional non-maximum suppression practice and adopting a two-stage model based on a gating cycle unit to integrate spatial and temporal features.

P. Zhang et al.[90] also chooses to train the character detector and group activity classifier end to end. This way shares the backbone network computing, improves the operational efficiency of the model, enables the character detector to detect critical figures, and improves the accuracy of group activity recognition. In addition, P. Zhang et al.[90] also proposed an interactivity relationship identification method based on weakly supervised learning, which excavates the interactivity relationship between people and groups through hidden vector learning and does not need the annotation data of interactivity relationship and individual activity.

Character detection and target tracking are essential in many group activity identification models. However, calculating these two steps is very time-consuming, and if character detection or target tracking needs to be corrected, it will interfere with the subsequent group activity identification. To solve this problem, Zhuang et al.[91] proposed a group activity recognition model without character detection and tracking. The model accepts video as input, directly uses the convolutional neural network to extract the global visual features of each video frame, and then uses the long and short-term memory network to integrate the multi-frame features and obtain the video representation for group activity classification.

**2.5 Relationship Model-based Methods**

Santoro et al.[92] first proposed the concept of a relation network. Figure 13 shows four objects accompanied by relational and non-relational questions. The former requires reasoning about object relations, while the latter pertains to attributes.

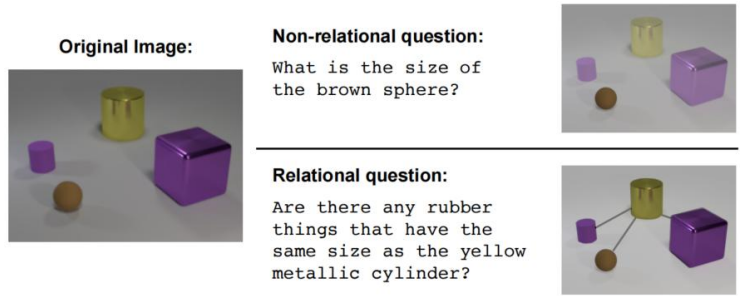

Figure 3 An illustrative example of relational reasoning[92].

Chowdhury et al.[93] proposed Hierarchical Relational Attention to extend RNs to Video QA. They achieved this by connecting attention modules over VGG and C3D features to hierarchical relational modules, which were nested according to the input question query token. Park and Kwak[94] introduced the RNs to 3D Pose Estimation. Their approach involved feeding estimated 2D joint coordinates into their proposed RN in a specific order, grouping them by body part to generate the estimated 3D position. Ibrahim et al.[33] utilized Hierarchical RNs for group activity recognition, starting from CNN features from each individual. Interactivity Relational Network(IRN) [16] is the first to extend RNs for Human Interactivity Recognition using pose information, as shown in Figure 4.

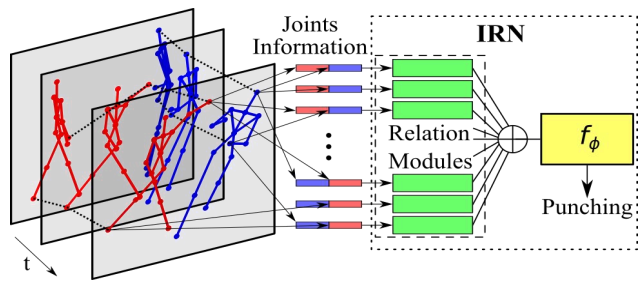

Figure 4 Interactivity Relation Network[16]

Since relational modeling is crucial for understanding multi-person scenarios and group activity, in the era of deep learning[95], relevant researchers try to model the interactivity between people and groups in multi-person systems by combining deep learning technology. Graph model, relationship network, and other technologies are applied to the group activity recognition model to extract the relationship information. Previous methods have generally used data such as RGB and optical flow in combination with neural networks like GCN, LSTM, and GRUs to recognize group activities, lacking the investigation of intergroup relationships. Based on IRN, Perez et al. made many extensions, such as fusion multiple relationship fusion, sequential relational reasoning, and domain knowledge pairwise structured operation. Then, Group Interactivity Relational Network(GIRN) [38], proposed by Perez et al., leverage coordinate information from human joints to construct multi-type relational modules. GIRN's algorithmic framework consists of multiple relational modules, each specializing in learning different relationships by processing skeletal information from each human body to gain information about those relationships. It can supplement behavioral information from various individuals to improve recognition. In addition, GIRN uses attention mechanisms to focus on the relationships of critical individuals. An overview of the GIRN method is shown in Figure 3.

To identify individual and group activities in multiplayer scenarios, Deng et al.[95] use neural networks to simulate the message-passing process on factor graphs, modeling the interdependence between different activity categories. Deng et al.[96] also tries to introduce the graph model into the deep network framework and use the rich relationship information in the scene to identify the group activity. The proposed structured inference framework uses a recurrent neural network to compute the messages transmitted between the nodes, and each node fuses the information sent by the surrounding nodes to update its state.

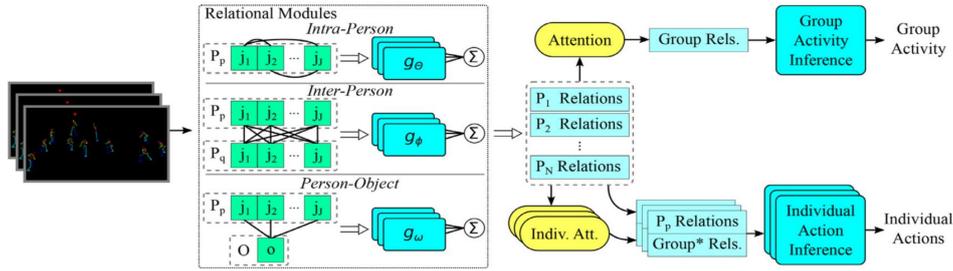

Figure 5 Overview of Group Interaction Relational Network (GIRN)[38]

Qi et al.[97] presents a neural network model for video group activity recognition using spatiotemporal context information. The model builds a semantic graph for each video, with nodes and edges denoting characters and their relationships. The model employs two recurrent neural networks to update the states of nodes and advantages according to their features and neighbors. The model also applies a spatiotemporal attention model to select the key characters and frames in the video. The model can learn the relationship between characters and extract distinctive spatiotemporal features.

Ibrahim & Mori[33] proposed a hierarchical relationship network model to calculate the relationship representation in the population. Given a graph structure and the characteristics of the potential connections between people, the relationship network layer computes the presentation of its relationship with adjacent nodes for each node. It obtains a new node representation using multiple relational network layers by superposition[92]. In addition, using different graph connection structures in each layer, we can gradually integrate individual character features to obtain global scene features for group activity identification. The author manually designed the relationship graph connection of each layer according to the spatial location relationship of the characters' structure. In addition, the author also proposed a relational autoencoder model to learn the character and scene representation unsupervised. The available terms can apply to activity and scene retrieval tasks using the K-nearest neighbor search algorithm, and the experimental results show that better retrieval results can achieve without annotated data. Azar et al.[98] designed a convolution relationship model using spatial relationships between individual characters to identify group activity. The model first generates activity activation maps for the input video image. The activity activation graph is an activity representation with a spatial structure. According to the position box of all characters, it assigns the possibility of different activity categories on each spatial region, including individual and group activity.

Group activity videos often contain a lot of characters and interactivity information, and then only a few characters in a few key frames determine the category of group activity. Therefore, it is necessary to efficiently model meaningful group relationships and suppress irrelevant individual activity and interactivity information to identify group activity accurately. Hu et al.[99] proposed a model based on deep reinforcement learning to gradually improve group activity's low-level characteristics and high-level relationships.

J. Wu et al.[100] also used the character relationship diagram to model the relationship information in the multi-person scene. The nodes on the graph represent the characteristics of each character, and the edges represent the relationship between the two characters. A learnable subnetwork calculates the shape relationship by taking the visual features of two characters and the shape weight between them as input. The position relationship is determined based on the spatial distance between the characters. The author designs and compares various methods of shape and spatial relations. The author proposes the integration method of multiple relationship graphs to enable the model to capture various related information in complex scenes. They use numerous subnetworks with different parameters to calculate various relationship graphs for an input video. On the constructed character relationship graph, the author uses the convolutional graph network[101] with the feature transfer on the chart. Each node obtains the relationship information according to the connection weight with other nodes.

D. Xu et al.[102] presented a group activity recognition model based on multimodality relationship representation. The model uses an object relationship module to model the appearance and location of objects and an optical flow extractive network to extract the motion of characters.

**2.6 Attention-based Methods**

In the human visual perception system context, certain input image regions hold more salience than others and likely exert a more significant influence on the final decision. Consequently, our brain tends to identify crucial areas of the image and allocate increased attention to them. Researchers have also attempted to apply the attention mechanism to neural networks in recent years for models to pay attention to crucial information in the input, discarding irrelevant noise and information, thereby improving model recognition accuracy[103].

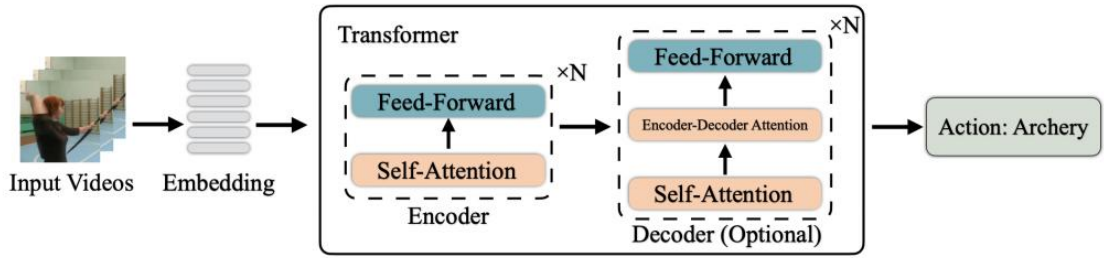

Figure 6 An illustration of Attention-based approaches on GAR

In the video of group activity in the multi-person scene, even though there is more than one character, only a tiny portion of the characters' actions determine the group activity category. Ramanathan et al.[104] proposed an attention mechanism-based model that can learn to detect critical players in determining group activity in scenarios involving multiple humans. Since the model does not require additional data, it can train exclusively on group activity data.

The visualization shows that characters with a high attention weight are often the key figures in the scene, such as the player firing. In addition,[104] presented a basketball game that defines activity events such as three-point shooting, dip shooting, and dunking. The proposed data set has more samples and volume than the previous group activity data set.

Yan et al.[105] also attempted to aid in accurately identifying group activity by locating key individuals in multiplayer scenarios. He proposes a model of attention for judging each character's participation at each moment and gradually incorporating the dynamic dynamics of the key characters. In this work, the author mainly considers the group activity participation of the characters through the movement characteristics of the characters and focuses on two types of key people: the first key person moves steadily in the whole video for a long distance; at one point, the second key person makes a rapid movement. The author believes these two types of characters will be the key to determining group activity in the multiplayer scene and should give a greater focus. Furthermore, we split the entire model into two parts: In Part 1, the network extracts the individual temporal characteristics of each character, and in Part 2, the network integrates the features of multiple characters for group activity classification based on the learned attention weights.

Tang et al.[106] proposed a Semantics-Preserving Teacher-Student (SPTS) model for recognizing the teacher's group activity, expecting the model to automatically discover critical characters in the scene through attention to semantic retention and discard irrelevant characters. He applies knowledge distillation to cause the model to focus on semantically significant individuals. The teacher network learns a group activity recognizer from the semantic text of the individual activities and assigns attention weights to each character. The student network learns from the teacher network and the input video with an attention mechanism. The student network mimics the teacher's attention weights and group activity scores. In this way, the student network preserves some of the semantics of the teacher network.

Tang et al. [107] built on SPTS [106] and added graph convolution modules in both teacher and student networks to strengthen SPTS. The graph represents the characters' features, with each node representing a character's feature and each edge representing the relationship between characters. The graph's structure models the dependency between them. It determines the adjacency matrix as a function of spatial position and uses the convolution network to transfer information across characters. He also used the intermediate features of the model to detect the temporal sequence of group activities and to locate the starting time point for the group activity of interest.

To model hierarchical interactivities in a group activity, Lu et al.[108] proposed a two-stage interactivity model of attention. At the level of the individual, it computes the attention weight between all characters in the scene at each time, and each individual obtains context information based on attention weights with other characters and updates its feature representation. Lu et al.[42] proposed an attention relation model of graphs, which can infer the interactivity relation in the scene from both the individual and group levels and learn the spatial and temporal evolution of group activity from these interactivity relations.

The transformer consists of one encoder and one decoder[109]. The encoder consists primarily of multiple self-attention blocks for encoding the input sequence. Except for an additional encoder-decoder attention mechanism in each block, the decoder shares a similar architecture with the encoder. This design enables the transformer to perform well concerning long-term dependency modeling, multimodality fusion, and multi-task processing. Figure 5 shows Attention-based activity recognition. A transformer [110] is a network structure based on the self-attention mechanism, first proposed in the machine translation task of natural language processing. Many research efforts in recent years have attempted to apply Transformer networks to computer vision tasks. Gavrilyuk et al.[43] leverage the Transformer model of characters selectively extracts information about the relationships between characters to identify group activities. For the input to the Transformer network to fully represent the appearance and motion characteristics of the characters, the author uses a 2D pose network and 3D convolutional neural network to extract static and dynamic representations for each character, respectively.

Ehsanpour et al.[111] focused on sub-group splitting in scenarios with multiple individuals. Multi-people methods can involve multiple social groups engaged in different social activities. In light of this, Ehsanpour et al. devised an end-to-end learnable model framework, which can partition characters into different sub-groups based on the social relationship between the characters and predict the individual activity of each nature and the social activity of each sub-group. The author uses the I3D network to extract each character's spatial and temporal characteristics and then uses the Transformer structure to improve the features of each character. The author then uses the graph-attention module to model individual interactivity. The model achieves good recognition performance on the traditional group activity identification data set with a single group. Furthermore, the author also performed sub-group splitting and sub-group activity annotation on the existing group activity identification dataset to verify the model's ability to identify multiple sub-groups in the scene.

# 3 Comparison of the-state-of-the-Technologies

This section will introduce group activity recognition relative datasets and the accuracy comparison of the state-of-the-technologies.

## 3.1 Datasets

Many data sets have been proposed to evaluate the performance of group relation activity recognition accurately approaches in recent years. Researchers collect most of these datasets from multiple-scene videos, such as sports competitions and security monitoring, where the labels for various group activities are defined. This section will introduce five popular datasets for interactivity activity recognition and provide information about their sources. Table 3 presents the basic information, including the data size, the number of categories, and the scenes from which we collected the videos. We taxonomy them as sports and surveillance according to the type of scene.

Table 3 Some Popular Datasets and Their Details.

| Dataset | Ref | Year | L- clips | Clips | Multi-types | Invid-types | Scene |
|---|---|---|---|---|---|---|---|
| NBA | [112] | 2007 | 55 | -- | 8 | 9 | Sport |
| Collective | [21] | 2009 | 44 | 2500 | 5 | 6 | Surveillance |
| UCLA Courtyard | [64] | 2012 | - | 106min | 6 | 10 | Surveillance |
| Volleyball | [5] | 2016 | 55 | 4830 | 8 | 9 | Sport |
| C-Sports | [113] | 2020 | - | 2187 | 5 | -- | Sport |

### 3.1.1 Sport dataset

The NBA dataset[112] was gathered from 181 NBA basketball games and contained a total of 9,172 video clips. The test set comprised 1,548 video clips, while the training set included 7,624. This dataset was only labeled with the group behavior tags of the videos but not with the individual activity tags of the people. There are 9 group behavior labels, including three types of activities: 2-point shot, 2-point layup, 3-point shot, and three types of activities: success, failure, and the offense get the rebound, loss, and the defense gets the rebound.

The Volleyball dataset [5] contains 4,830 short video clips collected from 55 volleyball matches, including 3,493 training sets and 1,337 test sets. Group activity categories are assigned to each video segment, encompassing eight labels: suitable set, right spike, right pass, correct pinpoint, left set, left end, left key, and left pinpoint. Furthermore, we annotate the middle frame of each video clip with character frames and movements. There are nine individual activity tags: waiting, setting, digging, failing, spiking, blocking, jumping, moving, and standing.

Unlike other motion scene datasets, the C-Sports dataset [113] contains only one movement. As part of the C-Sports dataset, Zalluhoglu C et al.[113]collected video data for 11 sports, including American football, basketball, dodgeball, handball, hockey, ice hockey, bag hockey, rugby, volleyball, and water polo. For each sport, we defined activities such as gathering, dismissal, passing, attack, and wandering. There are 2,187 video clips in the data set, including 1,317 training sets, 435 validation sets, and 435 test sets.

Researchers commonly utilize other sport scene datasets in group activity recognition studies, including NCAA[104], Sports-1M[114], MultiSports[115], and hockey[66].

**3.1.2 Surveillance dataset**

**The Collective dataset**[6] is surveillance scene data the camera collects. The surveillance camera video contains 44 sequences and about 2,500 keyframes, each as a sample. Generally, researchers select approximately one-third of the video sequences as the test set and choose the remaining two-thirds as the training set. The dataset has five group activity tags: crossing, waiting, queueing, walking, and talking, and six individual activity tags: NA, crossing, waiting, queueing, walking, and speaking. Each keyframe defines the most ongoing particular activity as a group activity.

**The UCLA Courtyard Dataset**[64] sources pictures collected from high-resolution surveillance cameras on campus, including 106 minutes of video. The authors defined six labels of group activity: Walking-together, Standing-in-line, Discussing-in-group, Sitting-together, Waiting-in-group, and Guided-tour. There are also ten individual character activity tags: Riding-skateboard, Riding- bike, Riding-scooter, Driving-car, Walking, Talking, Waiting, Reading, Eating, and Sitting. In addition, the author also marked the position and category of the 17 objects in the picture.

Additionally, Researchers widely use the BEHAVE[116], UCF Crime[117], NUS-HGA[19], and Nursing Home[118] as surveillance scene data set by researchers.

**3.1.3 Other**

There are many other datasets for group relation activity recognition, such as RGB-based, multi-modal-based, skeleton-based datasets, and so on. For RGB-based activity recognition, the HMD51[119], UCF101[120], Kinetics-400[121], Kinetics-600[122], Kinetics-700[123], Something-Something-V1[124], and Something-Something-V2[124] are popular benchmark datasets used by researchers. For multi-modal-based group relation activity recognition, the NTU RGB-D 60[125], NTU RGB + D 120[126], PKU-MMD[127], SDUFall[128], SBU Kinect[129], MSRActivity 3d[130], and NTU-MHAD[131] are widely used benchmark. For skeleton-based group relation activity recognition, the NTU RGB-D 60[125], NTU RGB + D 120[126], Northwestern-UCLA[132], UWA3D Multiview II[133] , and MSRDailyActivity3D[134] are widely used benchmark datasets. Researchers utilize point cloud-based group activity recognition datasets[135, 136] to evaluate the performance of point cloud human activity data.

However, no benchmark datasets exist for non-visual modality-based group relation activity recognition.

**3.2 Accuracy analysis**

The volleyball Dataset and Collective dataset are the most widely used in group activity recognition models in recent years. In this article, we chose these two datasets to evaluate the accuracy of group relation activity recognition of advanced mainstream methods. Table 4 presents the results, listing the techniques employed in each study. It demonstrates the performance comparison across various metrics and highlights the variations in performance among different method types.

Due to the differences in the feature extractive methods, backbone network models, and experimental settings used in different works, and the limitations of the data set itself. The accuracy rate can't fully reflect the advantages and disadvantages of the group activity recognition method. Based on the different performance comparisons, the following technical methods can improve the accuracy of the group activity recognition model.

1) Improve the characterization ability of the model input features. Regarding group activity identification, researchers have discovered many features that help distinguish different group activities, including visual features, spatial location structure features, relationship information features, scene context features, character posture features, timing dynamic features, etc. The model can more accurately identify group activity by introducing more input features and improving the accuracy of input features.

2) Relational modeling of the population. Group activity contains multi-relationship information: interactivity information between people, the relationship between humans and group, and the relationship between subgroup and subgroup. Modeling and using this relationship information can help the model to better distinguish between different group activities.

3) Use the attention mechanism to extract the critical information in the scene. The multiplayer group activity set is very complex, and there is much noise information unrelated to the activity category, which challenges the model recognition. The attention mechanism can help the model to focus on the critical information related to the group activity. What's more, it can suppress noise information. As can be seen from Table 4, the current approach is based on the attention mechanism[106, 137], leading to accuracy results obtained in both Collective and Volleyball.

4) Taking direction and size information into skeleton-based group relation activity recognition. Most researchers have chosen to model joint-level details for Skeleton-based interactivity relation activity recognition. However, the direction information of the skeleton is also crucial to inter-joints or intra-joints. People rarely consider a means of performing activity recognition with edge-level details, which may play a significant role in group relation activity

recognition[138]

Table 4 Accuracy Comparison with State-of-the-technologies mainly on Volleyball or Collective dataset.

| Method | Ref | Type | Accuracy(%) | |
|---|---|---|---|---|
| | | | Volleyball | Collective |
| HDTM | [5] | HRNN-based | 81.9 | 81.5 |
| AC | [22] | Descriptor-based | - | 68.2 |
| RSTV | [67] | Descriptor-based | - | 70.9 |
| VICAR | [25] | Descriptor e-based | - | 73.2 |
| MIR | [26] | Non-deep learning-based | - | 83.3 |
| Kernel | [27] | Non-deep learning-based | - | 83.4 |
| DCM | [139] | Non-deep learning-based | - | 85.5 |
| CRM | [140] | Deep learning-based | 93.0 | 85.8 |
| SBGAR | [141] | HRNN-based | 66.9 | 86.1 |
| CERN | [142] | HRNN-based | 83.3 | 87.2 |
| stagNet | [32] | Deep learning-based | 89.3 | 89.1 |
| HRN | [33] | Deep learning-based | 89.5 | - |
| SSU | [30] | HRNN-based | 90.6 | - |
| GAIM | [143] | Attention-based | 91.9 | 90.6 |
| ARG | [144] | Deep learning-based | 92.6 | 91.0 |
| Actor-transformers | [145] | Attention-based | 94.4 | 92.8 |
| SPA | [106] | Attention-based | 90.7 | 95.7 |
| Dual-AI | [37] | Relation model-based | 95.4 | 96.5 |
| GIRN | [38] | Relation model-based | 94.0 | - |
| STIGPN | [39] | Deep learning-based | 92.6 | 93.3 |

# 4 Existing Problems and Developing Prospects

The traditional interactivity activity recognition approaches suffer from drawbacks, including insufficiently detailed descriptions of interactive activities and the inability to leverage individual-level features fully. Although deep learning has succeeded in group activity recognition, there are still open research questions. In this section, we will talk about group activity recognition facing issues and future directions.

**4.1 Existing Problems**

According to this literature review, group activity recognition faces challenges as below:
1. Crowd-density unstructured populations increase the task challenges. To the best of our knowledge, as occlusions, illumination, motion, background, scale, camera angle, and complexity, current algorithms must be adequate for accurate identification under these complex conditions.
2. Inadequate consideration of internal links. Existing relational activity recognition approaches

rarely consider internal links, such as links between joint and unconnected nodes, motion direction, joint coordinates, etc.

3. Algorithm training or validation datasets are limited to a specific data set or data sets of a certain number. There is no single robust dataset for the training and validation of most algorithms, and there is a wide variety of datasets for using algorithms.

4. Complex algorithms need multiple GPUs and be more friendly to average or less-funded learners.

**4.2 Developing Prospects**

Group activity recognition technology has many applications in surveillance, sports, social occasions, and other tasks. However, the current approaches still need recognition accuracy, robustness, and running speed, which is difficult to large-scale applied in real scenarios. Here are some possible future research directions.

1) While the advanced models for group activity recognition that are widely adopted[41, 43] have achieved high identification accuracy on relevant datasets, the limited sample size of group activity identification datasets[5, 6] presents challenges in terms of generalizability and stability. As a result, the model may be prone to overfitting and need more strength in practical applications. In future work, we can start by expanding the scale of training data, enhancing the population representation ability of the model, optimizing the relationship modeling method, and suppressing noise information and other directions, to further improve the accuracy and robustness of the group activity identification model.

2) Construct accurate group activity detection, tracking, and recognition networks. Most work to date focuses primarily on the simple problem of group activity recognition, i.e., for an input short video segment, it is sufficient to output a group activity label. Real videos can have multiple groups to perform different group activities to detect and track. For now, the model must have group activity detection capability and can automatically localize the location of the spatial and temporal dimension of each group, the included characters, the ongoing group activity, and the start and end times of the action. Researchers have already conducted studies on detecting individual video activity[146, 147]. In contrast, there needs to be more video data sets and practical models for group activity detection problems in group activity research.

3) Integrate the group activity identification model with the person detection model and the tracking model. Many group activity identification methods rely on character detection and tracking results[5, 35, 41, 43]. However, most works use independent models to identify video detection, tracking, and group activity. Future work can try to study how to integrate these models. On the one hand, computing resources can be saved by sharing the underlying features; on the other hand, end-to-end learning can provide better results for group activity recognition through detection and tracking modules[30];

4) Improve the recognition accuracy of low-frequency group activity. In real group activity scenarios, some group activities occur at a shallow frequency, and accurately identifying these low-frequency abnormal activities is very useful, such as group fighting. The model's performance in these categories is poor because of the small samples of low-frequency population activity. Future work needs to improve the accuracy of model identification in the case of small sample markers and unbalanced category distribution.

# 5 Conclusion

We summarize the group activity recognition types based on the analysis of current research work and provide a detailed introduction to each method's contents and research progress. We present the background, application scenarios, and critical technical challenges. Then, we compare the recognition accuracy of advanced mainstream methods. Finally, we summarize the shortcomings of the current group activity recognition model and propose some possible research directions.

# Acknowledgments

We express our gratitude for the support provided by the Natural Science Foundation of Guangdong Provincial Higher Education Characteristic Innovation Research Foundation (Grant NO.2019KTSCX258) for this work.

# Declaration of competing interest

The authors assert that they have no identifiable competing financial interests or personal relationships that might have influenced the work presented in this paper.